\documentclass[]{spie}  

\usepackage[]{amssymb,amsmath, graphicx}

\title{Regression in Sensor Networks:  Training Distributively with Alternating Projections} 

\author{Joel B. Predd and Sanjeev R. Kulkarni and H. Vincent Poor
\skiplinehalf
Department of Electrical Engineering, Princeton University, Princeton, NJ, USA}

\authorinfo{E-mail:\{jpredd, kulkarni, poor\}@princeton.edu\\This research was supported in part by the Army Research Office under Grant
DAAD19-00-1-0466, in part by Draper Laboratory under Grant IR\&D
6002, in part by the National Science Foundation under Grants CCR-0020524 and 
CCR-0312413, and in part by the Office of Naval Research under Grant No.
N00014-03-1-0102. }
 
\begin{document} 
\def\IR{{\rm I \kern-0.20em R}}
\def\bbbr{{\IR}}
\newtheorem{thm}{Theorem}
\newtheorem{prop}{Proposition}
\newtheorem{cor}{Corollary}
\newtheorem{rem}{Remark}
\newtheorem{defn}{Definition}
\newtheorem{lem}{Lemma}
\newtheorem{conjecture}{Conjecture}
\newtheorem{fact}{Fact}
\newtheorem{example}{Example}
  \newcommand{\CITING}[1]{[\citenum{#1}]}

  \maketitle 

\begin{abstract}

Wireless sensor networks (WSNs) have attracted considerable attention in recent years and motivate a host of new challenges for distributed signal processing.  The problem of distributed or decentralized \emph{estimation} has often been considered in the context of parametric models.  However, the success of parametric methods is limited by the appropriateness of the strong statistical assumptions made by the models.  In this paper, a more flexible \emph{nonparametric} model for distributed regression is considered that is applicable in a variety of WSN applications including field estimation.  Here, starting with the standard regularized kernel least-squares estimator, a message-passing algorithm for distributed estimation in WSNs is derived. The algorithm can be viewed as an instantiation of the \emph{successive orthogonal projection} (SOP) algorithm. Various practical aspects of the algorithm are discussed and  several numerical simulations validate the potential of the approach.
\end{abstract}



\section{INTRODUCTION}
\label{sect:intro}  
\subsection{Motivation}
Wireless sensor networks have attracted considerable attention in
recent years \CITING{AkySuSanCay02}.  Research in this area has focused on two separate
aspects of such networks:  networking issues, such as capacity,
delay, and routing strategies; and applications issues.  This paper
is concerned with the second of these aspects of wireless sensor
networks, and in particular with the problem of distributed inference.
Wireless sensor networks are {\it a fortiori} designed for the purpose
of making inferences about  the environments that they are sensing,
and  they are  typically characterized by limited communications
capabilities due to tight energy and bandwidth limitations. Thus,
distributed inference is a major issue in the study of such networks.

The problem of distributed or decentralized \emph{estimation} has often been considered in the context of parametric models, in which strong assumptions are made about a statistical description of the environment.  In \CITING{DelNeeBar04}, for example,  the observations made by a sensor network are modeled by a Gaussian random field whose dependency structure is described by a graphical model;  in such a setting, iterative message passing algorithms such as belief propagation (BP) \CITING{Pea88}, loopy BP \CITING{WeiFre01}, embedded trees \CITING{SudWaiWil03}, embedded polygons \CITING{DelNeeBar04}, and Gaussian elimination \CITING{PlaKum02} can be used to efficiently estimate the field distributively.  \CITING{Now03} investigates a distributed EM algorithm for estimating a field modeled as a mixture of Gaussians.  In the case where sensors observe the same phenomenon through independent noisy channels, \CITING{BlaHer04} studies a distributed maximum likelihood estimator using an information sharing algorithm based on Fisher scoring.  \CITING{RabNow04,SonChiKulSch05} derive distributed algorithms based on incremental subgradient methods that are useful for parameter estimation in sensor networks. Finally, \CITING{RibGia04a} focuses on the case when sensor observations are quantized to one-bit.

As in the standard centralized case, the success of distributed parametric methods is limited by the appropriateness of the statistical assumptions made by the model.   In certain applications, such strong modeling assumptions are warranted and systems designed from these models show promise.  However, in other scenarios, prior knowledge is at best vague and translating such knowledge into a statistical model is undesirable.   Applications such as these pave the way for a nonparametric study of distributed estimation.  To this end, several nonparametric analyses have considered the distributed estimation problem. For example, weakening the assumptions made in \CITING{RibGia04a}, \CITING{Luo04} focuses on universal decentralized estimation in the case when sensor observations are quantized to one bit.  \CITING{PreKulPoo04a,PreKulPoo05a} study the existence of consistent estimators in several models for distributed learning.  


In this paper, motivated in part by the success of (reproducing) kernel methods in machine learning \CITING{SchSmo02}, we consider a nonparametric approach to distributed estimation that is based on regularized kernel least-squares regression. In particular, we assume that a network of sensors is distributed about a plane and that each sensor makes a local observation of a (say, temperature) field.  If every sensor's position and field measurement are available at a central fusion center, then the regularized least-squares estimator is a standard nonparametric estimate of the field (at locations perhaps not occupied by sensors).  More precisely, assuming that prior-knowledge about spatial correlations in the field can be encoded in a positive definite kernel, a standard centralized approach to field estimation is to minimize a regularized empirical loss functional over functions in the corresponding reproducing kernel Hilbert space \CITING{SchSmo02}.   

Of course, due to limitations on energy and bandwidth, the sensors' measurements and positions are not available at the fusion center and therefore, such centralized estimation schemes are not feasible in wireless sensor networks.  In this paper, however, we suggest a distributed algorithm for computing an approximation. To arrive at the approximation, we first interpret the classic estimator as a projection onto a Hilbert space and then consider a natural relaxation that is derived from the topology of the sensor network.  The relaxation suggests a \emph{local message-passing} algorithm based on the \emph{SOP algorithm} (successive orthogonal projection)\CITING{CenZen97} for distributively  estimating the field.  Through several numerical experiments, it is observed that the algorithm converges after a reasonable number of iterations,  that the accuracy of the distributed estimate is improved through the local message-passing algorithm, and that the estimate is comparable with the centralized kernel least-squares estimator.  

\subsection{Related Work}
The work presented here is most similar to the research presented in \CITING{GueBodThiPasMad04}, where a general framework for distributed kernel linear regression is explored.  The network model is essentially identical to the one considered there and thus, it is useful to contrast the methods.  First, whereas the current work is cast formally within the context of reproducing kernel Hilbert spaces, \CITING{GueBodThiPasMad04} takes a weight-space view and attempts to estimate the field as a linear combination of local basis functions. Secondly, in \CITING{GueBodThiPasMad04}, the algorithm is based on a distributed implementation of Gaussian elimination using junction trees.  In the current approach, a message-passing algorithm is derived from the SOP algorithm and a network topology dependent relaxation of the centralized least-squares estimator.   Finally, in this paper, regularization is incorporated into the model.  

Alternating projection algorithms are often applied in other areas of statistical signal processing, for example, restricted least-squares regression \CITING{Dyk83} and computerized tomography \CITING{Sta87}.  In \CITING{HerBla05}, an alternating projection approach was taken to the problem of localization in sensor networks.  However, to our knowledge, such techniques have not been applied more generally to the problem of distributed, communication-constrained estimation. 

Alternating projection methods have often been used in algorithms for parallel optimization \CITING{CenZen97}. Algorithms similar to those presented in this paper may be useful to help circumvent the complexity induced by massive data sets in machine learning, possibly by parallelizing kernel methods.   To our knowledge, SOP (or other alternating projection algorithms) have not been applied in this related context.

Given the connection between kernel methods and Gaussian processes \CITING{Wah00}, those familiar with message-passing algorithms such a belief-propagation may find the algorithm presented here familiar.  Formalizing such a connection would likely require one to interpret our ``relaxation" in the context of dependency structures in Gaussian processes, and to connect alternating projection algorithms with the generalized distributive law \CITING{AjiMce00}.  

\subsection{Organization}
The remainder of the paper is organized as follows.  In Section 2, we introduce notation and then review the SOP algorithm and regularized least-squares regression in reproducing kernel Hilbert spaces.  In Section 3, we formalize the sensor network model and derive an algorithm for distributed estimation in sensor networks.   In Section 4, we summarize several experiments with simulated data where the algorithms are evaluated.  Finally, conclusions, extensions, and future work are discussed in Section 6.

\section{PRELIMINARIES}


\subsection{Alternating Projections Algorithms}
Let $\cal{X}$ be a Hilbert space with a norm denoted by $\|\cdot\|$.  Let $C_1, \ldots, C_n$ be closed convex subsets of $\cal{X}$ with a nonempty intersection $C = \cap_{i=1}^n C_i$.  Let $P_{C}$ denote the orthogonal projection of ${\hat{x}}\in {\cal{X}}$ onto $C$, i.e.,
\begin{eqnarray}
\nonumber P_{C}( \hat{x}) \triangleq \arg \min_{x\in C} \| x - \hat{x}\|.
\end{eqnarray}
Define $P_{C_i}$ analogously. The following question arises:  can $P_C(\cdot)$ be computed using $\{P_{C_i}(\cdot)\}_{i=1}^n$?  Iteratively defined through (\ref{SOP}), the (unrelaxed) SOP (successive orthogonal projection) algorithm is a natural approach:
\begin{eqnarray}\label{SOP}
x_{0}:=\hat{x} & x_{n} := P_{C_{n\mod{ m + 1}} }(x_{n-1}).
\end{eqnarray}
In words, SOP successively and iteratively projects onto each of the subsets; much of the behavior of this algorithm can be understood through the following lemma, whose proof can be found in \CITING{BauBor96} (see Remark 6.19).  
\begin{lemma}
Let $\{C_i\}_{i=1}^m$ and $C$ be as above.  Let $x_n$ be defined as in (\ref{SOP}). Then,  for every $x\in C$ and every $n$,
\begin{equation}
\nonumber\|x_n -x\| \leq \|x_{n-1} - x\|.
\end{equation}
Moreoever, $\lim_{n\rightarrow\infty} x_n \in \cap_{i=1}^m C_i$.  If $C_i$ are subspaces for all $i\in\{1,...,m\}$, then $\lim_{n\rightarrow\infty} \|x_n - P_C(\hat{x})\| = 0$.  If the angle between the sets is strictly positive, then convergence is linear.
\end{lemma}

This algorithm is often studied in the context of the \emph{convex feasibility problem} and has a wide-range of signal processing applications (e.g., image reconstruction).  Having been generalized in various ways \CITING{CenZen97}, SOP often takes on other names (e.g., Bregman's algorithm).  Below, we comment on how these generalizations may be useful in practical settings.  For the sake of space, we forego a thorough examination of alternating projection algorithms and their applications and refer the reader to standard references on the topic \CITING{BauBor96,CenZen97}.


%

\subsection{Regularized Kernel Least-squares Regression}
Reproducing kernel Hilbert spaces are often considered in statistical signal processing and have been popularized within the machine learning community by the success of kernel methods.  In this section, we briefly review regularized kernel least-squares regression in order to anchor our notation.  For a more thorough introduction, we refer the reader to various references on the topic; see, for example, \CITING{KaiPoo98,SchSmo02} and references therein.

Let $X$ and $Y$ be ${\cal{X}}$ and ${\cal{Y}}$-valued random variables, respectively.  ${\cal{X}}$ is known as the feature, input, or observation space; ${\cal{Y}}$ is known as the label, output, or target space.  In the sequel, we take ${\cal{X}}=\IR^2$ to model the positions of sensors in the plane and we take ${\cal{Y}}= \IR$ to model real-valued sensor field measurements.   In the least-squares estimation problem, we seek a decision rule mapping inputs to outputs that minimizes the expected squared error.  In particular, we seek a function $g:{\cal{X}}\rightarrow {\cal{Y}}$  that minimizes
\begin{equation}
{\mathbf{E}}\{|g(X) - Y|^2\}.
\end{equation}
It is well-known that $\eta(x) = {\mathbf{E}}\{Y\,|X=x\}$ is the loss minimizing rule.  However, without prior knowledge of the joint distribution of $(X,Y)$, this regression function cannot be computed.  In the supervised learning problem, one is instead provided a sequence of input/output \emph{training examples}, $S=\{(x_i, y_i)\}_{i=1}^n\subset{\cal{X}}\times{\cal{Y}}$; the learning task is to use $S$ to estimate $\eta(x)$.  Regularized kernel least-squares methods offer one approach to this problem.

In particular, let ${\cal{H}}_K$ denote the \emph{reproducing kernel Hilbert space} (RKHS) induced by a \emph{positive semi-definite kernel} $K(\cdot, \cdot):{\cal{X}}\times{\cal{X}}\rightarrow\IR$.  More precisely, we can associate with any positive definite function $K(\cdot, \cdot)$ a unique collection of functions ${\cal{H}}_K$ such that
\begin{eqnarray}
\nonumber K_t = K(\cdot, t) &\in&{\cal{H}}_K \,\,\, \forall t\in{\cal{X}}\\
\label{RKHS-linearity} \sum_{i=1}^m \alpha_{i} K_{t_i}&\in&{\cal{H}}_K \,\,\,    \forall\{\alpha_i\}_{i=1}^m\subset\IR,\,\, m<\infty.
\end{eqnarray}
If we equip ${\cal{H}}_K$ with an inner-product defined by $<K_s, K_t>  = K(s, t)$, extend ${\cal{H}}_K$ using linearity  to all functions of the form (\ref{RKHS-linearity}), and include the point-wise limits, then ${\cal{H}}_K$ is called an RKHS.  Note that
\begin{equation}
\nonumber <K_x, f> = f(x),
\end{equation}
for all $x\in{\cal{X}}$ and all $f\in{\cal{H}}_K$; this identity is the reproducing property from which the name is derived.  Finally, let $\|\cdot\|_{{\cal{H}}_K}$ denote the norm associated with ${\cal{H}}_K$.

\begin{example}
When ${\cal{X}}=\IR^d$, a canonical RKHS is the one associated with the linear kernel, $K({\mathbf{x}}, {\mathbf{x}}^{\prime}) = {\mathbf{x}}^T {\mathbf{x}}^{\prime}$.  \end{example}
\begin{example}
When ${\cal{X}}=\IR^d$, a second canonical RKHS is the one associated with the Gaussian kernel, $K({\mathbf{x}}, {\mathbf{x}}^{\prime}) = \exp^{-\|{\mathbf{x}} - {\mathbf{x}}^{\prime}\|_2^2}$.  \end{example}

In the numerical studies in Section 4, we will explore examples using both the linear and Gaussian kernels.

Returning to our discussion of least-squares estimation, let us apply this RKHS formalism to the regression problem.  Given a positive semi-definite kernel $K$, which is often designed as a similarity measure for inputs, the regularized kernel least-squares estimate is defined as the solution $f_{\lambda}\in{\cal{H}}_K$ of the following optimization problem:
\begin{eqnarray}
\label{kernel}\min_{f\in{\cal{H}}_K}\Big{[} \sum_{i=1}^n (f(x_i) - y_i)^2  + \lambda \| f \|_{{\cal{H}}_K}^2\Big{]}.
\end{eqnarray}
The first term in objective function (\ref{kernel}) is a measurement of how well the rule ``fits the data"; the second term acts as a complexity control.  $\lambda\in\IR$ is a constant parameter that governs the trade-off between these two terms.  In the learning context, the statistical behavior of this estimator is well-understood under various assumptions on the stochastic process that generates the examples $\{(x_i, y_i)\}_{i=1}^n$ \CITING{GyoKohKrzWal02,SchSmo02}. Moreover, this highly successful technique has been verified empirically in applications ranging from bioinformatics to hand-written digit recognition. 

For those unfamiliar with kernel methods, let us note in passing that this estimator has a useful variational interpretation in the context of Gaussian process estimation \CITING{Wah00}.  For example,  one may posit a statistical model such as
\begin{equation}
\nonumber y_i = \mu(x_i) +\epsilon_i,
\end{equation}
and assume that $(\mu(x))_{x\in{\cal{X}}}$ is zero-mean Gaussian process with covariance function $k(x, x^{\prime})$ and that $\{\epsilon_i\}_{i=1}^n$ is independent and identically distributed (i.i.d.) zero-mean Gaussian noise.  Under these assumptions, $f_{\lambda}(x) ={\mathbf{E}}\{\mu(x)\,|\,S\}$.

In this paper, we focus on algorithmic aspects of computing a solution to (\ref{kernel})  (or an approximation thereof) in distributed, communication-constrained environments like sensor networks.  To this end, consider the following ``Representer Theorem" proved originally in \CITING{KimWah71}.  The result is significant because it states that while the objective function (\ref{kernel}) is defined over a potentially infinite dimensional Hilbert space, its minimizer must lie in a finite dimensional subspace.
\begin{theorem}[\CITING{KimWah71}]
Let $f_{\lambda} \in{\cal{H}}_K$ be the minimizer of (\ref{kernel}).  Then, there exists ${\mathbf{c}}_{\lambda}\in\IR^n$ such that
\begin{equation}
\label{representer} f_{\lambda}(\cdot) = \sum_{i=1}^n c_{\lambda,i} K(\cdot, x_i).
\end{equation}
\end{theorem}
After plugging (\ref{representer}) into (\ref{kernel}), one can derive that 
\begin{equation}
\label{solution}{\mathbf{c}}_{\lambda} = (K + \lambda I)^{-1}{\mathbf{y}},
\end{equation}
where $K = (k_{ij})$ is the kernel matrix ($k_{ij} = K(x_i, x_j)$).  In practice, the matrix inversion in (\ref{solution}) is often avoided by using algorithms for solving linear systems; often, sparsity structure in $K$ can be exploited.  In the learning literature, the process of computing ${\mathbf{c}}_{\lambda}$ given $S$ is often called {\emph{training}}, whereas \emph{testing} generally refers to the process evaluating the field estimate $f_{\lambda}(x) = \sum_{i=1}^n c_{\lambda,i} K(x, x_i)$  for a new input $x$.

Finally, note that (\ref{kernel}) can be interpreted as an orthogonal projection.  In particular, by introducing an auxiliary vector ${\mathbf{z}}\in\IR^n$, (\ref{kernel}) can be rewritten as the following optimization program:
\begin{eqnarray}
\label{rlsqreg-r1}\min & \| {\mathbf{z}} - {\mathbf{y}}\|_2^2 + \lambda \| f \|_{{\cal{H}}_K}^2\\
\label{extraconstraints}{\textrm{s.t.}} & z_i = f(x_i) &\forall i\in\{1,...,n\}\\
\nonumber & {\mathbf{z}}\in{\IR^n}\\
\nonumber & f\in{\cal{H}}_K.
\end{eqnarray}
Through the constraints in (\ref{extraconstraints}), the optimization problems are equivalent in the following sense:  if $f_{\lambda}$ is the minimizer of (\ref{kernel}) and $({\mathbf{z}}^{\prime}, f_{\lambda}^{\prime})$ is the solution of (\ref{rlsqreg-r1}), then $f_{\lambda}^{\prime} = f_{\lambda}$.  Therefore, through (\ref{rlsqreg-r1}), we can interpret the regularized kernel least-squares estimator as a projection of the vector $({\mathbf{y}}, 0)\in\IR^n\times{\cal{H}}_K$ onto the set 
\begin{equation}\nonumber
\Big{\{}({\mathbf{z}}, f)\in\IR^n\times{\cal{H}}_K \, : \, z_i = f(x_i) \,\,\forall i\in\{1,...,n\}\Big{\}}\subset\IR^n\times{\cal{H}}_K.
\end{equation}
This simple observation will recur in the sequel.

\section{DISTRIBUTED LEAST-SQUARES REGRESSION}

\subsection{The Model}
Consider a network of $n$ sensors that are distributed about the plane $\IR^2$;  let $\{{\mathbf{x}}_i\}_{i=1}^n\subset\IR^2$ denote the coordinates of the sensors' positions.   Assume that each sensor can accurately localize itself in the plane, i.e., assume that for all $i\in\{1,..., n\}$, sensor $i$ knows ${\mathbf{x}}_i$ ; this assumption is justified by the existence of various localization algorithms for wireless sensor networks (e.g., \CITING{HerBla05}).  In what follows, we will interchangeably refer to the $i^{th}$ sensor in the network by its index, $i$, or its position, ${\mathbf{x}}_i$.

Suppose that the sensors form a wireless ad-hoc network with a topology described by a graph; for example, consider the topology depicted in Figure 1.  Each node in the graph represents a sensor in the network; an edge between two nodes posits the existence of a point-to-point (wireless) communication link between the corresponding two sensors.  Let $N_i\subseteq\{1,...,n\}$ denote the set of neighbors for sensor $i$.  Interpret $N_i$ as the set of sensors with which sensor $i$ can communicate directly.  Consistent with this interpretation, let us abuse notation slightly and assume that $i\in N_i$ for every $i\in\{1,...,n\}$;  in other words, each sensor can communicate with itself.    We make no additional assumptions on the structure of the graph  (i.e., we do not require it to be connected nor to form a spanning tree, etc.).  However, the performance of the algorithms will indeed be effected by such properties and we comment on such relationships below.

\begin{figure}
   \begin{center}
   \begin{tabular}{c}
   \includegraphics[height=4cm]{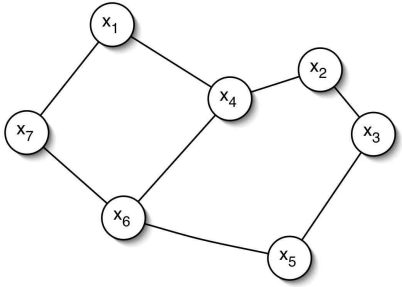}
   \end{tabular}
   \end{center}
   \caption[Network] {Sample Sensor Network Topology}
 \end{figure} 
Note that in this model, the capacity of communication links in the network (i.e., edges in the graph) is not modeled explicitly.  Instead,  we assume that each link can support the simple messages to be passed by our algorithm; this assumption is consistent with other applications of message-passing algorithms (e.g., belief-propagation) to sensor networks.

After the network has been localized, assume that each sensor makes a noisy estimate of  a (say, temperature) field at its position; let $y_i$ denote the measurement of sensor $i$.  One may posit a relationship between the sensor's measurement and its position.  For example, one could assume that
\begin{equation}
y_i  = {\textrm{temperature}}({\mathbf{x}}_i) + n_i,
\end{equation}
for some Gaussian random field $({\textrm{temperature}}({\mathbf{x}}))_{{\mathbf{x}}\in\IR^2}$, and an i.i.d. zero-mean noise sequence $\{n_i\}_{i=1}^n$.  However, such assumptions are not required in this model for nonparametric regression.

Assume that a centralized fusion center wishes to use the sensors' local measurements to infer the field across the plane.  In other words, the fusion center would like to determine a function $f:\IR^2\rightarrow\IR$ such that the expected squared error,
\begin{equation}\nonumber
{\mathbf{E}}\{ | f(X) - {\textrm{temperature(X)}}|^2\},
\end{equation}
is small.  Here, the expectation could be with taken respect to any probability distribution defined on $\IR^2$; for instance, the distribution may model the mechanism through which sensors are randomly positioned in the plane.

If the fusion center had access to every sensor's position and temperature measurements, the regularized least-squares estimator is a natural choice.  Using the formalism of Section 2.2, one may take ${\cal{X}} = \IR^2$ to model the positions of sensors in the plane and take ${\cal{Y}}=\IR$ to model the sensors' real-valued field measurements. After choosing a kernel $K(\cdot, \cdot)$ to reflect the designer's prior knowledge of spatial correlation in the field, the fusion center may estimate the field as $f_{\lambda}$, the solution to 
\begin{eqnarray}
\label{rlsqreg}\min_{f\in{{\cal{H}}_K}} \Big{[}\sum_{i=1}^n (f({\mathbf{x}}_i) - y_i)^2  + \lambda \| f \|_{{\cal{H}}_K}^2\Big{]}.
\end{eqnarray}
Here, as in Section 2.2, ${\cal{H}}_K$ denotes the RKHS generated by the kernel $K$.  Of course, due to constraints on energy and bandwidth, the fusion center will not have access to any of the data in $\{({\mathbf{x}}_i, y_i)\}_{i=1}^n$ and therefore (\ref{rlsqreg}) cannot be solved using centralized methods; in general, classical kernel methods are infeasible for the sensor network.  Moreover, even if an algorithm exists to solve (\ref{rlsqreg}) distributively, the sensors may not be able to share the estimated model (i.e., $f_{\lambda}$) with the fusion center, also due to the complexity of communicating.  In short, our assumption is that the tight communication constraints preclude both centralized training and centralized testing in this model for distributed regression.

In what follows, we show how the optimization problem in (\ref{rlsqreg}) can be relaxed for the problem of distributed regression.   The relaxation will be derived in light of the topology of the sensor network and will suggest an algorithm for distributively estimating the field.   In the algorithm, each sensor will communicate only with neighboring sensors and each sensor will locally determine its own global estimate for the field.  In addition, we will suggest several algorithms for aggregating the sensors' estimates at the fusion center.

\subsection{An Algorithm for Distributed Regression}
To begin, let us associate a function $f_i\in{{\cal{H}}_K}$ and a parameter $\lambda_i\in\IR$ with sensor $i$ .  Now, consider the following optimization program:
\begin{eqnarray}
\label{rlsqreg-r2}\min_{} &\sum_{i=1}^n (f_i({\mathbf{x}}_i) - y_i)^2  + \sum_{i=1}^n \lambda_i \| f_i \|_{{\cal{H}}_K}^2\\
\label{coupling1}{\textrm{s.t.}} & f_j({\mathbf{x}}_i) = f_i({\mathbf{x}}_i) & \forall i,j\in\{1,...,n\}\\
 \nonumber & f_i\in{{\cal{H}}_K} & \forall i\in\{1,...,n\}
\end{eqnarray}
Here, the optimization variables are $\{f_i\}_{i=1}^n\subset {{\cal{H}}_K}$; $\{({\mathbf{x}}_i, y_i)\}_{i=1}^n\subset\IR^2\times\IR$ and $\{\lambda_i\}_{i=1}^n\subset\IR$ are the program data.  The \emph{coupling constraints} in (\ref{coupling1}) dictate that for any feasible solution to (\ref{rlsqreg-r2}), every sensor's associated function is equivalent when evaluated at $\{{\mathbf{x}}_i\}_{i=1}^n$.    As a result, one can think about (\ref{rlsqreg-r2}) as an equivalent form of (\ref{rlsqreg}) in the following sense.

\begin{lemma}
Let $(f_{\lambda_1},...,f_{\lambda_n})\in{{\cal{H}}^n_K}$ denote the solution of (\ref{rlsqreg-r2}) and let $f_{\lambda}\in{{\cal{H}}_K}$ denote the solution of (\ref{rlsqreg}).  Assume that $\lambda_i>0\,\,\forall i\in\{1,...,n\}$. Then, $f_{\lambda_1} = \cdots = f_{\lambda_n}$.  If $\sum_{i=1}^n \lambda_i = \lambda$, then $f_{\lambda} = f_{\lambda_1}$.
\end{lemma}

This form of the regularized least-squares regression problem suggests a natural relaxation that allows us to incorporate the communication model of the sensor network into the estimator.  In particular, we relax the coupling constraints as follows: 
\begin{eqnarray}
\label{rlsqreg-r4}\min_{} &\sum_{i=1}^n (f_i({\mathbf{x}}_i) - y_i)^2  + \sum_{i=1}^n \lambda_i \| f_i \|_{{\cal{H}}_K}^2\\
\label{coupling2} {\textrm{s.t.}}& f_j({\mathbf{x}}_i) = f_i({\mathbf{x}}_i) & \forall j\in N_i, \forall i\in\{1,...,n\}\\
\nonumber & f_i\in{{\cal{H}}_K} & \forall i\in\{1,...,n\}
\end{eqnarray}
In comparison to (\ref{rlsqreg-r2}), the coupling constraints in (\ref{rlsqreg-r4}) require two sensors' functions to be equivalent only when evaluated at shared neighbors' locations.  More precisely, for any feasible solution to (\ref{rlsqreg-r4}), $f_i({\mathbf{x}}_k) = f_j({\mathbf{x}}_k)$ if $k\in N_i\cap N_j$.  Since we have assumed $i\in N_i$ for all $i\in\{1,\ldots, n\}$, this interpretation is equivalent to the constraints stated formally in (\ref{coupling2}).

Next, note that just as (\ref{rlsqreg}) can be interpreted as a projection via (\ref{rlsqreg-r1}), (\ref{rlsqreg-r4}) can be interpreted as a (weighted) projection of the vector $({\mathbf{y}}, 0, \ldots, 0)\in\IR^n\times{{\cal{H}}^n_K}$ onto the set $C = \cap_{i=1}^n C_i$, with
\begin{equation}
C_i = \Big{\{}({\mathbf{z}}, f_1, \ldots, f_n)\,:\, {\mathbf{z}}\in\IR^n, f_i\in{{\cal{H}}_K}, {\textrm{\,\,and\,\,}} f_i({\mathbf{x}}_j) = z_j \,\,\forall j\in N_i\Big{\}}\subset\IR^n\times{{\cal{H}}^n_K}.
\end{equation}
This fact is straightforward to derive using the same manipulation applied in Section 2.2.   The significance of this observation lies in the fact that the relaxed form of the regularized kernel least-squares estimator has been expressed as a projection onto the intersection of a collection of $n$ convex sets; in particular, note that each set $C_i$ is a subspace.  Thus, the SOP algorithm can be used to solve the relaxed problem (\ref{rlsqreg-r4}).  Moreover, computing $P_{C_i}(\cdot)$ requires sensor $i$ to gather information only from its neighbors in the network.  More precisely, note that for any ${\mathbf{v}}=({\mathbf{z}}, f_1, \ldots, f_n)\in\IR^n\times{{\cal{H}}^n_K}$, $P_{C_i}({\mathbf{v}}) = ({\mathbf{z}}^{\star}, f_1^{\star}, \ldots, f_n^{\star})$ where
\begin{eqnarray}
\nonumber f_j^{\star} & = & f_j \,\,\,\,\,\, \forall j\neq i\\
\nonumber f^{\star}_i &=& \arg\min_{f\in{{\cal{H}}_K}} \sum_{j\in N_i} (f({\mathbf{x}}_i) - z_i)^2 + \lambda_i \| f - f_i\|_{{\cal{H}}_K}^2\\
\nonumber z_j^{\star} & = & z_j   \,\,\,\,\,\,\forall j\notin N_i\\
\nonumber z_j^{\star} & = & f_i^{\star}({\mathbf{x}}_j)  \,\,\,\,\,\,\forall j\in N_i
\end{eqnarray}
To emphasize, computing $P_{C_i}({\mathbf{v}})$ leaves $z_j$  unchanged for all $j\notin N_i$ and leaves $f_j$ unchanged for all $j\neq i$. The function associated with sensor $i$, $f_{i}^{\star}$ can be computed using $\{z_j\}_{j\in N_i}$ and $f_i$.  Thus, tying these observations together, we are left with an algorithm for distributively estimating the field by solving a relaxed form of the regularized least-squares estimator (\ref{rlsqreg-r4}).  The algorithm, SN-Train, is summarized in psuedo-code in Table 2 and depicted pictorially in Figure 2.

\begin{table}[htdp]
\begin{center}
\begin{tabular}{|ll|}
\hline
\textbf{Initialization:} & Kernel update: sensors notified which kernel to use.\\
& Sensor localization: sensor $s$ determines ${\mathbf{x}}_s$\\
& Neighbors share locations: sensor $s$ stores $\{{\mathbf{x}}_j\}_{j\in N_s}$\\
& Sensor $s$ initializes $z_{s,0} = y_i$, $f_{s,0} = 0\in{{\cal{H}}_K}$ \\
&\\
\textbf{Train:} & for $t=1,\ldots, T$ \\
&   \hspace{1cm}for $s=1,\ldots, n$ \\
&  \hspace{2cm}Sensor $s$: \\
&  \hspace{2.5cm}Queries $z_{j, t-1}\,\,\forall j\in N_s$\\
&  \hspace{2.5cm}$f_{s, t} := \arg\min_{f\in{{\cal{H}}_K}} \Big{[}\sum_{j\in N_s} (f({\mathbf{x}}_j) - z_{j, t-1})^2 + \lambda_s \| f  - f_{s, t-1}\|_{{\cal{H}}_K}^2\Big{]}$  \\
& \hspace{2.5cm}Updates $z_{j, t} = f_{s, t}({\mathbf{x}}_j)\,\,\,\forall j\in N_s$\\
& \hspace{1cm} {\textrm{end}}\\
&  {\textrm{end}}\\
\hline
\end{tabular}
\end{center}
\label{default}
\caption{SN-Train}
\end{table}%

\begin{figure}
   \begin{center}
   \begin{tabular}{c}
   \includegraphics[height=8cm]{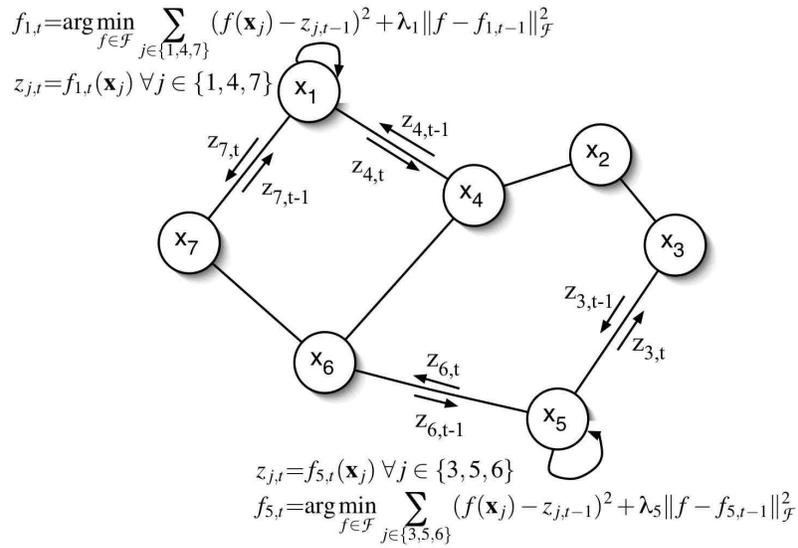}
   \end{tabular}
   \end{center}
   \caption[SN-Train] {Depiction of Training phase of SN-Train}
 \end{figure} 
Before proceeding with a discussion of the algorithm, let us briefly mention two important properties.  First, note that the asymptotic behavior of SN-Train is implied by the analysis of the SOP algorithm.  In particular,  we have the following.

\begin{lemma}
Let $(f_{\lambda_1},\ldots, f_{\lambda_N})\in{{\cal{H}}^n_K}$ be the solution to (\ref{rlsqreg-r4}) and let $\{f_{s,T}\}_{s=1}^n\subset {{\cal{H}}_K}$ be as defined in SN-Train. Then,
\begin{equation}
\lim_{T\rightarrow\infty} f_{s,T} = f_{\lambda_s}
\end{equation}
for all $s\in\{1,\ldots, n\}$.
\end{lemma}

This lemma follows from Lemma 2.1 and the fact that convergence in norm implies point-wise convergence in an RKHS.  Given the structure of RKHS and the general analysis in \CITING{BauBor96}, SN-Train is expected to converge linearly for many kernels.  We forego a discussion of this important, but technical point for the sake of space.

Observe that Lemma 3.2 characterizes the output of SN-Train relative to (\ref{rlsqreg-r4}).  This characterization is useful insofar as it sheds light on the relationship between the output of SN-Train and (\ref{rlsqreg}), the centralized regularized least-squares estimator.  The following straightforward generalization of Theorem 2.2 is a step toward further understanding this important relationship.

\begin{lemma}
Let $(f_{\lambda_1},\ldots, f_{\lambda_n})\in{{\cal{H}}^n_K}$ be the solution to (\ref{rlsqreg-r4}) .  Then, for every sensor $s\in\{1,\ldots,n\}$, there exists ${\mathbf{c}}_{\lambda_s}\in\IR^{|N_s|}$ such that
\begin{equation}\label{newrepresent}
f_{\lambda_s}(\cdot) = \sum_{i\in N_s} c_{\lambda_s,i} K(\cdot, {\mathbf{x}}_i).
\end{equation}
\end{lemma}

The proof of this lemma follows from the original  Representer Theorem (applied to the update equation for $f_{s, t}$) and the fact that ${\cal{H}}^n_K$ is closed.  The significance of the result lies in the fact that connectivity fundamentally limits the accuracy of any sensor's global estimate of the field.  In particular, a sensor connected to few sensors in an otherwise dense network will be limited to estimates that lie in the span of functions determined by its neighbors;  thus, connectivity influences the bias of the network.  Intuitively, however, local message-passing in SN-Train should aid in optimizing the estimator within that limited span as long as the network is connected;  this intuition is borne out in experiments in Section 4.  

\subsection{Discussion}
Now, let us consider several issues relevant to the practical implementation of SN-Train.

\begin{description}
\item[Communication] In the training phase of the SN-Train algorithm, communication occurs only between neighboring sensors during the \emph{Query} and \emph{Update} stages of the iterative process.   In the \emph{Query} stage, the message from sensor $i$ to sensor $j$ is an estimate of the field at sensor $i$; in the \emph{Update} stage, the message from sensor $i$ to sensor $j$ is sensor $i$'s ``new" estimate of the field at sensor $j$.  To emphasize, though each sensor retains a locally estimated function, the messages passed through the network are \emph{not functions}, but real numbers that represent the network's estimate of the field at the sensor locations.  Moreover, though only local information is exchanged, global information is conveyed.   In particular, there are examples of RKHSs paired with very sparse network topologies for which the output of SN-Train is equivalent to the centralized estimator.

\item[Computation] In training phase of the SN-Train, computation occurs when sensor $s$ updates $f_{s,t}$.  By a mild generalization of Theorem 2.2 \CITING{SchSmo02}, ${f_{s,t}}(\cdot) = \sum_{j\in N_s} c_{s, t, j} K(\cdot, {\mathbf{x}}_j)$ for some vector ${\mathbf{c}}_{s,t}\in\IR^{|N_s|}$.  By substituting this representer into the update equation for $f_{s,t}$ in SN-Train, the update rule can be derived in terms of  ${\mathbf{c}}_{s, t}$ as follows:
\begin{equation}
\label{cupdate}{\mathbf{c}}_{s,t} = (K_s + \lambda_s I)^{-1}({\mathbf{z}}_{s,t-1} + \lambda_s {\mathbf{c}}_{s, t-1}),
\end{equation}  
where ${\mathbf{z}}_{s,t-1}\in\IR^{|N_s|}$ is appropriately defined to be the vector of measurements that sensor $s$ receives during the \emph{Query} stage.  Note that in principle, sensor $s$ can compute ${\mathbf{c}}_{s,t}$  efficiently if the number of neighbors is reasonably small, as is anticipated in sensor networks.  Also note, assuming that the field is slowly varying (i.e., ``neighboring sensor observations are correlated") and the sensor network topology is ``local", then the kernel matrix $K_s$ is not likely to be sparse.

\item[Parallelism] As described in Table 1, the inner loop of SN-Train iterates over the sensors in the network serially.  Note this ordering is non-essential and parallelism may be introduced.  In fact, two sensors can train simultaneously as long as they do not share an neighbors in the network (as depicted in Figure 2).  In practical settings, various multiple-access algorithms (e.g., ALOHA) may be adapted to negotiate an ordering in a distributed fashion.  Since the SOP algorithm and Lemma 2.1 have been generalized for a very general class of (perhaps random) control orderings \CITING{BauBor96}, Lemma 3.2 can be extended in many cases.

\item[Localization] In the initialization phase of SN-Train, sensors are assumed to store the positions of neighboring sensors. What is actually crucial  is that each sensor $s$ stores an evaluation of its local kernel matrix $K_s = [k_{{ij}}]_{ij\in N_s}\in\IR^{|N_s|\times|N_s|}$, which is needed at each update of ${f_{s,t}}$ (see the comments on Computation, above).  However, since many popular kernels depend only on the distance between sample points (e.g., $k_{ij} = \exp^{-\|{\mathbf{x}}_i - {\mathbf{x}}_j\|_2^2}$ for the Gaussian kernel), localization may be overkill.  It may be sufficient for each sensor to estimate relative distances between all neighboring sensors.

\item[Storage]  As mentioned in the comments on localization, it is crucial that each sensor $s$ stores an evaluation of its local kernel matrix $K_s$. Moreover, each sensor must store a representation of $f_{s,t}$.  By (\ref{cupdate}), it is sufficient to store an additional $|N_s|$ coefficients. Thus, the storage requirement of a sensor is influenced entirely by how many neighbors it has in the network.  In short, sensor $s$ must maintain $O(|N_s|^2)$ floating point numbers.

\item[Centralized special case] Note that it may be useful to think of this model for distributed regression as a strict generalization of the centralized kernel linear regression framework.  In particular, the standard kernel linear regression model corresponds to a fully connected network in the sense described by Lemma 3.1.

\item[Robustness]  The SN-Train algorithm can be made robust to a changing network topology, for example due to sensor failures, etc.   For the sake of space, we will not elaborate the point here, but note in passing that SN-Train can be adapted to allow the neighborhood $N_{s,t}$ of sensor $s$ to be a function of time.  The resulting algorithm is essentially identical to SN-Train (allowing for necessary localization information to be exchanged midway).  Moreover, the algorithm converges to the solution implied by the largest stationary neighborhood that occurs ``infinitely often" and in a well-defined sense, ``progress" is made at each iteration (even when the ``current" neighborhood does not correspond to the largest).  An analysis of this robust approach requires several of the SOP generalizations discussed in \CITING{BauBor96}.

\item[Aggregation] As discussed above, it may be unreasonable to assume that the sensors can share the output of the distributed training algorithm, $\{f_{s, T}\}_{s=1}^n$, due to the complexity of communication in wireless sensor networks.  For the sake of the subsequent experimental studies, we will assume that when the fusion center wishes to find the temperature at ${\mathbf{x}}\in\IR^2$, it may employ one of the following three strategies to aggregate the response of the network.

\begin{itemize}
\item \textbf{Single Sensor} The fusion center will simply choose one sensor arbitrarily, say sensor $s$, and use it for all future field estimates.  Note that such a rule is feasible, since in SN-Train, each sensor locally derives global estimate $f_{s,T}$ for the field.

\item \textbf{$k$ Nearest-neighbor} The fusion center will average the estimates provided by the $k$ sensors nearest ${\mathbf{x}}$.  If $\sigma\subseteq\{1,...,n\}$ is the set  of the $k$ sensors nearest ${\mathbf{x}}$, then fusion rule will aggregate according to the following rule:
\begin{equation}
f_{k-\textrm{NN}}({\mathbf{x}}) = \frac{1}{k} \sum_{s\in\sigma} f_{s, T} ({\mathbf{x}}).
\end{equation}
Both the simple \emph{nearest-neighbor} and simple \emph{network average} are special-cases of the $k$ nearest-neighbor fusion rule, corresponding to $k=1$ and $k=n$, respectively.

\item \textbf{Connectivity-averaged} The fusion center take a weighted average of the estimates provided by all $n$ sensors.  The weights in the average will be determined by how connected the sensors are in the network.  In particular,  the fusion center will consider the following aggregation strategy:
\begin{equation}
f_{CA}({\mathbf{x}}) = \frac{1}{\sum_{s=1}^n |N_s|} \sum_{s=1}^n |N_s| f_{s, T} ({\mathbf{x}}).
\end{equation}

\end{itemize}
\end{description}

\section{Experiments}
In this section, we study SN-Train numerically through several simulations.  Indeed, each of the issues discussed in Section 4 requires further examination.  However, due to limited space, we focus our attention.  First, we study the convergence rate of SN-Train.  In particular, we study the number of outer iterations (i.e., $T$) needed for the expected error of fusion center's estimate (when using the various fusion rules discussed in Section 4) to converge.   Since even local communication can be a burden on power and delay if the number of necessary iterations is excessive, we hope to quantify this aspect of our iterative approach.  In a second study, we examine the quality of the estimate derived through SN-Train.   Since any algorithm for distributed estimation is ultimately validated by the accuracy of the estimate,  we examine the effect that local communication has on estimation quality.

\subsection{Setup}
In our model description in Section 3, we have assumed that sensors are distributed physically in the plane. However, the algorithm and results hold for any Euclidean\footnote{Actually, since kernel methods can be derived for \emph{any} input space for which a reproducing kernel Hilbert space embedding can be defined, our approach is broadly applicable.} space (e.g., $\IR^d$ for $d\neq 2$).  Thus, to simplify our study, we take $d=1$ and assume that $n$ sensors are distributed uniformly along the interval ${\cal{X}} = [-1, 1] \subset \IR$.   

In what follows, we model the observations made by the sensors as,
\begin{equation}
y_i = \eta(x_i) + n_i,
\end{equation}
where $\{n_i\}_{i=1}^n$ is an i.i.d. sequence of zero-mean Gaussian random variables with variance $\alpha^2$.  We consider two cases that differ primarily in the way we choose $\eta(\cdot)$.  To keep the problems interesting, we also modify $\alpha$ and $K$ (the kernel employed by the sensors) according to our selection of $\eta(\cdot)$.  In particular, we consider the following two cases.

\textbf{Case 1:} $\eta(x) = \eta_1(x) = 5x + 5$,  $\alpha=7$,  $K$ = linear kernel.

\textbf{Case 2:} $\eta(x) = \eta_2(x) = \sin(\pi x)$,  $\alpha=1$, $K$ = Gaussian kernel.

For convenience, Case 1 and Case 2 are contrasted pictorially in Figure 3.  A random sampling of $n=50$ sensor positions is depicted along the lower horizontal border of each plot; a corresponding sampling of sensor observations is plotted as well.  In Case 1, the regression function is linear and hence, local information is useful in constructing global field estimates.   In Case 2, the regression function has been constructed so that local information is comparatively less useful in constructing global field estimates.  As discussed below, this difference will influence the performance of SN-Train in various ways.  In what follows, we examine both convergence and estimation quality in experiments using Case 1 and Case 2.
\begin{figure}
   \begin{center}
   \begin{tabular}{c}
   \includegraphics[height=6.5cm]{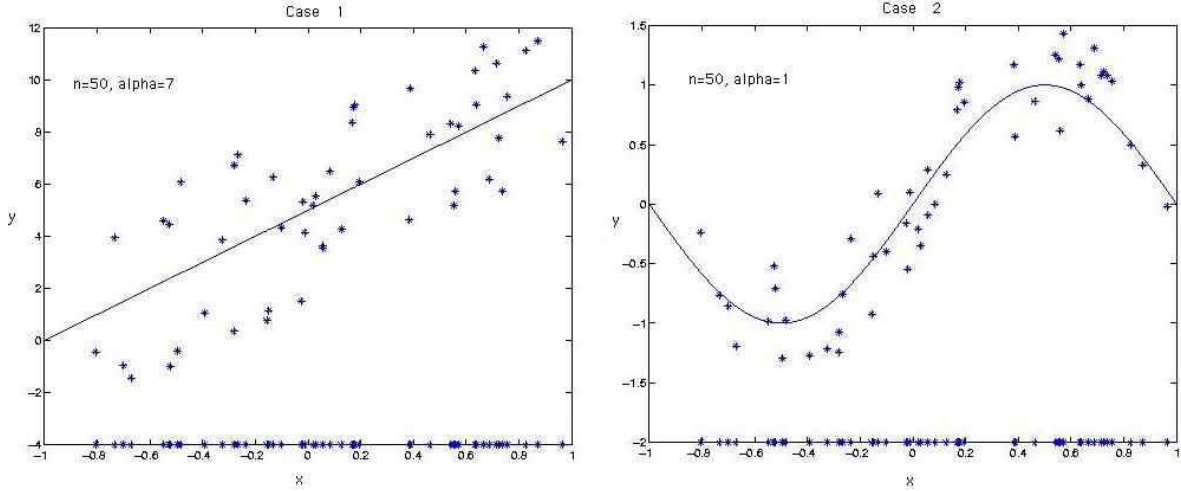}
   \end{tabular}
   \end{center}
   \caption[Convergence] {Two Base Cases}
 \end{figure} 
The parameter $r\geq 0$ is used to define the network topology for a random sensor network.  In particular, two sensors are neighbors in the network if the (Euclidean) distance between them is less than $r$.  Finally, somewhat arbitrarily, we select the regularization parameters so that $\lambda_i = \frac{\kappa}{|N_i|^2}$, with $\kappa = 0.01$ for all $i\in\{1,\ldots,n\}$.  

\subsection{Convergence}
To investigate the convergence rate of SN-Train, we vary $T$ from $1$ to $100$.  For each $T$, the regression function is randomly sampled $500$ times; the test error of the single sensor, nearest neighbor, and connectivity-averaged fusion rule is measured on this held out test set after the network has been trained distributively with SN-Train.  The results for Case 1 and Case 2 are plotted in Figures 4 and 5, respectively.  On the left, we randomize over $S=200$ random sensor positions and noise sequences and plot the average test error vs. $T$;  on the right, a plot a typical output estimate under each of the fusion rules.  For perspective, the error rate of the standard centralized kernel least-squares estimator is plotted as well.  Note the error rate plots are logarithmic on the dependent axis.  

Observe that for both Case 1 and Case 2, convergence is reasonably fast for each of the three fusion rules. In particular, note that the nearest-neighbor rule converges after two to three outer iterations.  Further, note that the nearest-neighbor fusion rule ultimately out-performs the other aggregation strategies, and is competitive with the centralized estimator.  Moreover, the single sensor fusion rule is quite poor; despite the fact that each sensor retains a global field estimate, the bias of the estimate is constrained by Lemma 3.  It interesting to note that the single sensor fusion is more competitive in Case 1 than in Case 2.  Since for linear rules, global information is useful for local estimates, individual sensors benefit from receiving local messages that are derived from distant sensors.  

We note in passing that these observations are fundamentally tied to the simulated relationship between the network topology and the kernels chosen.  In particular, in these experiments, neighboring sensors can communicate \emph{and} can expect to have correlated (i.e., similar) field measurements.  For instance, the convergence rate is not as fast and the nearest neighbor fusion rule is in general not superior when a ``random" network topology is paired with a Gaussian kernel.  In the interest of space, we do not explore this important point in this paper.
\begin{figure}
   \begin{center}
   \begin{tabular}{c}
   \includegraphics[height=6.5cm]{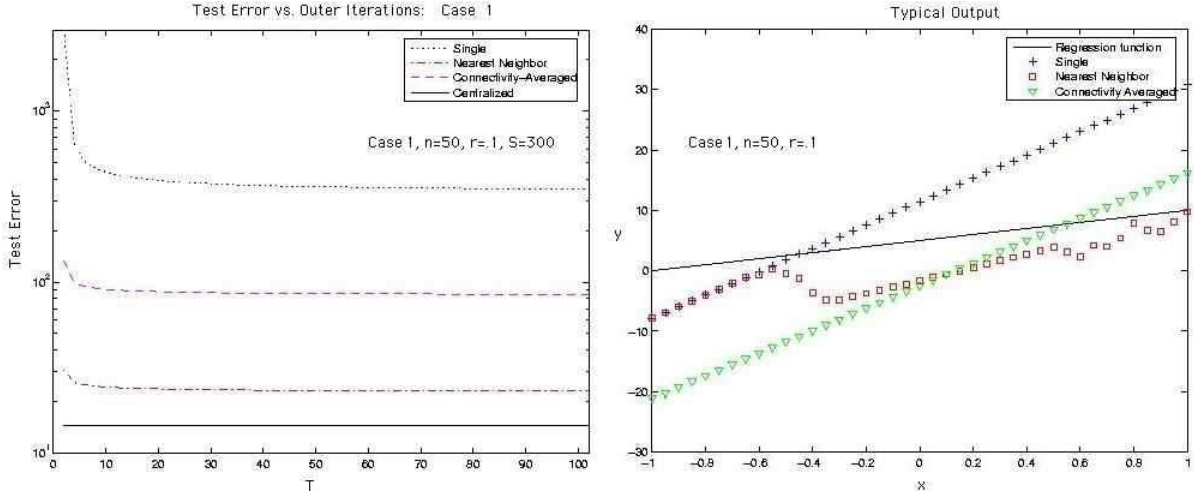}
   \end{tabular}
   \end{center}
   \caption[Convergence] {Convergence: Case 1}
 \end{figure} 

\begin{figure}
   \begin{center}
   \begin{tabular}{c}
   \includegraphics[height=6.5cm]{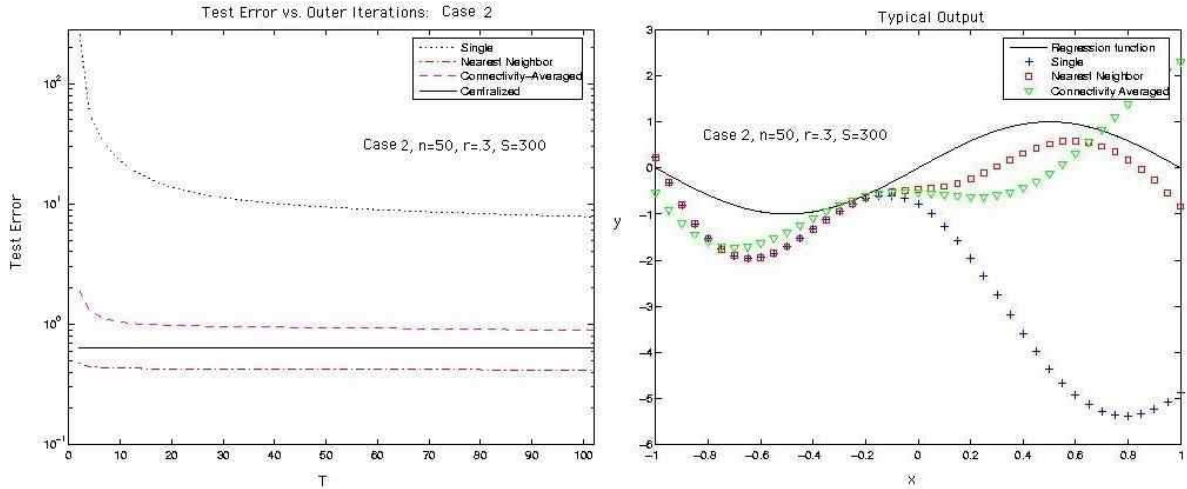}
   \end{tabular}
   \end{center}
   \caption[Convergence] {Convergence: Case 2}
 \end{figure} 
\subsection{Communication}
Since SN-Train adds complexity to the estimation algorithm, it is important to verify that this expense buys improved estimation accuracy.  To do so, we contrast training distributively with SN-Train with a ``local only" approach.  In the local only approach, we make only one iteration through the network, avoiding the \emph{Update} step in SN-Train.  That is, each sensor computes a global estimate for the field using the samples observed in its neighborhood;  however, information about these estimates is not exchanged through the message variables.  Since the only difference between using SN-Train and ``local only" is the Update step, the value of communication has been isolated.  For perspective, we also construct the standard centralized least-squares estimator estimator (with $\lambda=\frac{.01}{n^2}$) and plot its error rate which is independent of $r$. (i.e., we assume all the samples are centrally located).

In these experiments, we select $T=200$ to ensure that SN-Train has converged.  The estimation accuracy is measured through the expected squared error;  in particular, we randomly sample the regression function $300$ times and measure the test error empirically on this held out test set.    We study the test error at various levels of connectivity of the network.  In Case 1, we vary $r$ from $0.1$ to $0.6$ in increments of $0.05$ and in Case 2, we vary $r$ from $0.1$ to $2.1$ in increments of $0.1$.  For each $r$, we randomize over $S=300$ random sensor positions and noise sequences, and plot the average test error versus $r$ for SN-Train, local only, and the centralized approach.  We focus on the single sensor fusion rule, which allows us to compare how the local message passing influences each sensor's global estimate.  

Intuitively, as $r$ grows, we expect the error of local only approach to decrease, since each sensor has a greater number of neighbors.  The bias of the estimate should decrease since by Lemma 3, the sensors' estimates lie in the span of a larger set of basis functions; the variance should decrease since a greater number of samples are incorporated to average out the noise.  A priori, such trends are not expected for SN-Train.  

The results for Cases 1 and 2 are plotted in Figure 6.   Note, for clarity, we plot the test error in Case 2 logarithmically\footnote{This operation exacerbates the variance of our estimate; whereas the test error for the centralized approach should be independent of $r$, variance is observed in Figure 6.}.  For Case 1, our expectations are confirmed for the local only approach;  the test error decreases with increased connectivity.  Moreover, the estimate derived from SN-Train  also improves as connectivity increases.  For Case 2, the test error decreases with connectivity for SN-Train, but not for the local only approach.  We attribute this behavior to the fact that we are studying the single sensor fusion rule.  For low levels of connectivity, individual sensors have very little information that is useful for estimating the field outside their neighborhood.  Thus, for the sinusoidal regression function, the local only approach incurs large test error when its performance is assessed ``globally". Interestingly, the local communication in SN-Train significantly improves over this naive approach; while both techniques are fundamentally constrained by Lemma 3, the local message passing in SN-Train leads to a better estimate for each fixed level of connectivity.  Further, note that after a certain degree of connectivity, the estimate derived from SN-Train is competitive with the centralized approach. The level of connectivity required for the local only approach to be similarly competitive is larger for Case 2 as compared to Case 1; for Case 2, the network is nearly fully connected before the local only approach is competitive with the centralized estimator.
%

\begin{figure}
   \begin{center}
   \begin{tabular}{c}
   \includegraphics[height=6.5cm]{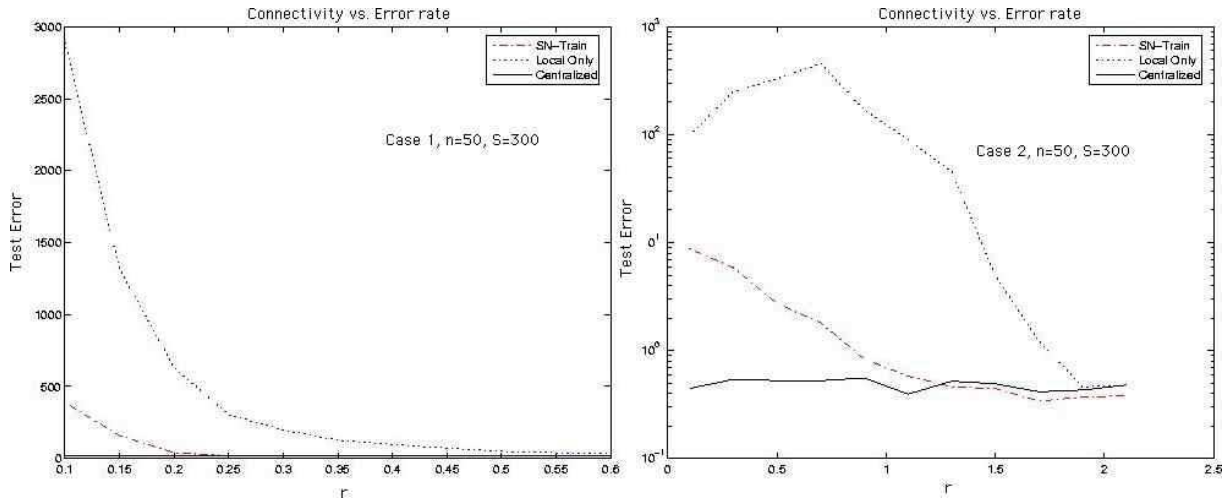}
   \end{tabular}
   \end{center}
   \caption[Connectivity] {Connectivity: Case 1 and Case 2}
 \end{figure} 

\section{Conclusions, Extensions, and Future Work}

\subsection{Conclusions}
There are several contributions of the current paper to the literature on distribution estimation. First, though kernel methods were advocated in \CITING{Sim03} and studied in the context of distributed detection in \CITING{NguWaiJor04}, to our knowledge regularized reproducing kernel methods have not been previously explored in the context of distributed estimation. Second, alternating projection algorithms are added to the list of iterative algorithms that can be usefully deployed for problems of distributed inference in sensor networks. Finally, a host of new statistical questions arise about the estimator, to be formalized in future work.  

\subsection{Extensions}
The SOP algorithm has been generalized to a wide variety of non-orthogonal projections including \emph{Bregman projections} \CITING{CenZen97}.  Thus, using this generalization, similar algorithms can be developed for distributed estimation in sensor networks for other loss functions (i.e., non-squared error) and more general regularizers.  In particular, the algorithms in the current paper can be generalized to handle loss functions and regularizers specified by \emph{Bregman divergences} \CITING{CenZen97}.  Moreover, as discussed in Section 3.3, SN-Train can be generalized in various ways to allow parallel and robust sensor update rules.

\subsection{Future work}
Future work will consider the extensions mentioned above and will analyze the statistical behavior of SN-Train.  In particular, we focus on understanding the design of regularization parameters that yield stable and consistent estimates.  Such work will rely on the extensive literature on the statistics of centralized kernel regression, as well as on various analyses of alternating projection methods.

\acknowledgments     
 Many of the ideas in this paper were inspired by the authors' ongoing collaboration \CITING{PreOshKulPoo05b} with Daniel Osherson on the problem of aggregating human expertise  \CITING{OshVar05}.  

\small
\bibliography{PreKulPoo05-SPIE05}   
\bibliographystyle{spiebib}   

\end{document}